\documentclass[letterpaper, 10 pt, conference]{ieeetran}  

\IEEEoverridecommandlockouts                              

\usepackage{graphics} 
\usepackage{epsfig} 
\usepackage{mathptmx} 
\usepackage{times} 
\usepackage{amsmath} 
\usepackage{amssymb}  
\usepackage[
colorlinks=true,
linkcolor=blue,
citecolor=blue,
filecolor=red,
urlcolor=blue]{hyperref}
\usepackage{algorithm}
\usepackage{eucal}
\usepackage[usenames,dvipsnames]{xcolor}
\usepackage{algpseudocode}
\usepackage{verbatim}
\usepackage{tabularx}
\usepackage{mathrsfs}
\usepackage{cite}
\usepackage{balance}
\usepackage{booktabs} 
\usepackage{tabularx}
\usepackage{color}
\algrenewcommand\algorithmicrequire{\textbf{Input:}}
\algrenewcommand\algorithmicensure{\textbf{Output:}}
\algrenewcommand{\algorithmiccomment}[1]{\hfill{\footnotesize\textcolor[gray]{0.2}{$\triangleright$ #1}}}

\newcommand{\xxnote}[3]{}
\ifx\hidenotes\undefined
  \renewcommand{\xxnote}[3]{\color{#2}{#1: #3}}
\fi

\title{\LARGE \bf
PRoID: Predicted Rate of Information Delivery in \\ Multi-Robot Exploration and Relaying
}

\author{Seungchan Kim$^{1{\dagger}}$, Seungjae Baek$^{2}$, Micah Corah$^{3}$, Graeme Best$^{4}$, Brady Moon$^{5}$, Sebastian Scherer$^{1}$
\thanks{This work was supported by Defense Science and Technology Agency (DSTA) under Contract \#DST000EC124000205.}
\thanks{$^{1}$ Carnegie Mellon University Robotics Institute}
\thanks{$^{2}$ Ulsan National Institute of Science \& Technology}
\thanks{$^{3}$ Colorado School of Mines}
\thanks{$^{4}$ University of Technology Sydney}
\thanks{$^{5}$ Brigham Young University}
\thanks{$^{\dagger}$ Corresponding Author: \texttt{seungch2@andrew.cmu.edu}}}

\begin{document}

\maketitle
\thispagestyle{empty}
\pagestyle{empty}

\begin{abstract}
We address Multi-Robot Exploration and Relaying (MRER): a team of robots must explore an unknown environment and deliver acquired information to a fixed base station within a mission time limit.
The central challenge is deciding when each robot should stop exploring and relay: this depends on what the robot is likely to find ahead, what information it uniquely holds, and whether immediate or future delivery is more valuable.
Prior approaches either ignore the reporting requirement entirely or rely on fixed-schedule relay strategies that cannot adapt to environment structure, team composition, or mission progress.
We introduce \textbf{\texttt{PRoID}} (Predicted Rate of Information Delivery), a relay criterion that uses learned map prediction to estimate each robot's future information gain along its planned path, accounting for what teammates are already relaying.
\textbf{\texttt{PRoID}} triggers relay when immediate return yields higher information delivery per unit time.
We further propose \textbf{\texttt{PRoID-Safe}}, a failure-aware extension that incorporates robot survival probability into the relay criterion, naturally biasing decisions toward earlier relay as failure risk grows.
We evaluate on real-world indoor floor plan datasets and show that \textbf{\texttt{PRoID}} and \textbf{\texttt{PRoID-Safe}} outperform fixed-schedule baselines, with stronger relative gains in failure scenarios.

\end{abstract}

\section{INTRODUCTION}
Deploying teams of robots to explore unknown environments has broad applications in robotics, including search and rescue \cite{queralta2020collaborative} and infrastructure inspection \cite{cabrita2010infrastructure}.
While most early exploration approaches optimize for raw map coverage \cite{yamauchi1998frontier, burgard2005coordinated}, practical deployments share a stricter requirement: acquired information must reach a fixed base station within a limited mission time \cite{de2011using}.
A robot that has mapped an entire building but fails to return before the deadline provides no operational value to the human operators waiting at the base.
This motivates a formulation distinct from standard exploration, where the objective is to maximize the information that actually reaches the base station, not merely the accumulated information that robots have observed.

At the core of this setting lies a fundamental tradeoff: how often each robot should return to relay.
Frequent relaying delivers data reliably but sacrifices exploration coverage; exploring until the deadline maximizes coverage but risks catastrophic data loss if the robot fails before returning.
In a multi-robot team, this tradeoff becomes richer: rather than the binary choice of relay-now-or-later, a robot can offload relay duty to a teammate and continue exploring.
This flexibility introduces new complexity--each robot must reason not only about its own exploration progress and unique information, but also whether a teammate is better positioned to deliver it.

\begin{figure}[t!]
    \centering
\includegraphics[width=1.0\linewidth,trim=1mm 5mm 120mm 0mm, clip]{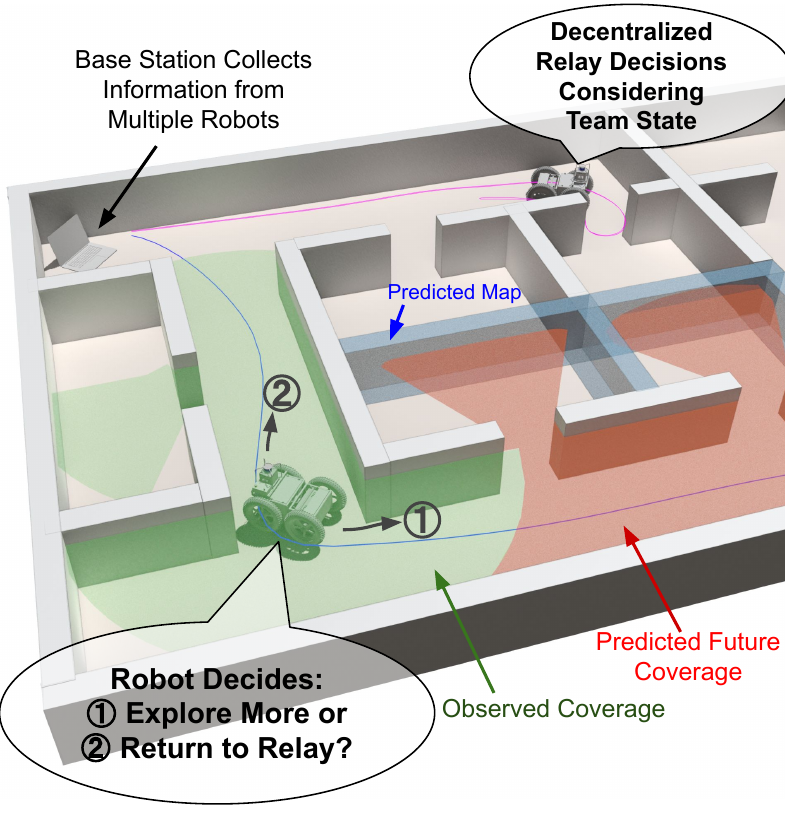}
    \caption{Multi-robot exploration must not only cover unknown environments but also deliver collected information to a base station in time. \textbf{\texttt{PRoID}} enables adaptive relay decisions by comparing each robot's current observed coverage against its predicted future sensor coverage along the planned path.}
    \label{Fig1}
\end{figure}

Prior approaches address parts of this problem but do not resolve the core challenge.
Frontier-based multi-robot exploration methods coordinate coverage efficiently \cite{yamauchi1998frontier, burgard2005coordinated}, but treat all observed data as immediately useful and ignore the reporting requirement entirely.
Rendezvous-based methods \cite{meghjani2011combining, song2024multi} focus on inter-robot information sharing rather than relay to a fixed base station, and their reliance on fixed schedules or preplanned meeting locations limits applicability in unknown environments.
Methods that do address the relay decision more directly, however, all rely on fixed, non-adaptive criteria.
Periodic return strategies \cite{hollinger2010multi,hollinger2012multirobot} relay at fixed intervals regardless of the environment structure or the team state.
Final relay strategies \cite{scherer2022resilient} maximize exploration but are brittle under failure: a robot that fails before returning loses all accumulated data.
Queue-based approaches \cite{clark2021queue} using fixed information thresholds fail to account for how relay overhead scales with environment size.
None of these approaches provides an adaptive, principled criterion for online relay decisions that considers predicted future gains and team coordination state.

Our key insight is that the relay decision can be framed as a comparison of information delivery rates: whether returning to the base now or exploring further first yields more information delivered per unit time.
We introduce the \textit{Predicted} Rate of Information Delivery (\textbf{\texttt{PRoID}}), which uses learned map prediction \cite{shrestha2019learned, ramakrishnan2020occupancy,ho2025mapex} to estimate the future information gain a robot is likely to collect along its planned path, while also considering what teammates are already relaying.
\textbf{\texttt{PRoID}} compares this predicted future delivery rate against the current delivery rate and triggers relay when returning immediately is more efficient  (Fig.~\ref{Fig1}). This allows the robot to adapt online to its state and the predicted environment structure without any pre-planned schedule or rendezvous.

We further consider the practically important setting in which robots may fail during the mission, permanently losing any information they have not yet delivered.
Fixed-schedule strategies cannot adapt relay decisions to a robot's growing failure risk, making them brittle in this setting.
We propose \textbf{\texttt{PRoID-Safe}}, a failure-aware extension of \textbf{\texttt{PRoID}} that incorporates the robot's survival probability into the relay criterion, naturally adapting its relay decisions to protect accumulated information as failure risk grows.

In summary, our contributions are:
\begin{itemize}
    \item We formalize the Multi-Robot Exploration and Relaying (MRER) problem that explicitly incorporates base station reporting requirements and robot failure probability.
    \item We introduce \textbf{\texttt{PRoID}} (Predicted Rate of Information Delivery), a relay criterion that uses learned map prediction to estimate each robot's future information gain along its planned path, accounting for what teammates are already relaying, and triggers adaptive online decisions without pre-planned schedules or rendezvous.
    \item We propose \textbf{\texttt{PRoID-Safe}}, a failure-aware extension of \textbf{\texttt{PRoID}} that incorporates survival probability into the relay criterion, remaining robust in failure scenarios where fixed-schedule baselines degrade significantly.
    \item We demonstrate through simulation on real-world indoor floor plans that \textbf{\texttt{PRoID}} and \textbf{\texttt{PRoID-Safe}} outperform fixed-schedule baselines, with particularly strong improvements in failure-prone scenarios.
\end{itemize}

\section{Related Works}

\subsection{Multi-Robot Exploration}

Multi-robot exploration has been extensively studied to cover large unknown environments efficiently. Early work assigned targets based on utility functions weighted by distance and expected coverage \cite{yamauchi1998frontier, burgard2005coordinated}, or by segmenting the environments into regions for allocation \cite{wurm2008coordinated}. Later approaches used geometric features \cite{kim2023multi}, learned coordination policies \cite{chiun2025marvel}, or suboptimality-bounded sequential greedy assignments \cite{corah2017efficient} to improve coverage efficiency. However, all these methods treat exploration as the end goal: coverage achieved is assumed immediately useful, with no model of the requirement that information must reach a base station within a time limit. Our work adopts multi-robot frontier-based exploration as the underlying coverage mechanism, but introduces a relay criterion that these methods lack.

\subsection{Relay Decision-Making Strategies}
Robots exploring unknown environments while acquiring and sharing information with teammates and a base station have been studied extensively. A more comprehensive review of the field is summarized in the survey \cite{amigoni2018multirobot}. One line of work addresses this by assigning \textit{fixed roles to robots}: explorers and relays \cite{de2009role, de2011using, goddemeier2012role}. Relays traverse the environment to collect information from explorers and return to the base station. Although this role-specialized strategy maintains connectivity effectively, it sacrifices exploration capacity since only a subset of the team actively contributes to coverage. While \cite{pei2010coordinated} proposes dynamic per-iteration role assignment, it still requires centralized computation at the base station. Our approach instead enables flexible, fully decentralized role assignment based on predicted information gain.

An alternative class of approach allows \textit{every robot to both explore and relay}—the formulation we adopt. Many existing systems use fixed or predefined relay strategies: periodic reconnection \cite{hollinger2010multi, hollinger2012multirobot} is robust but not adaptive to environment structure or teammates' progress; final-only strategies \cite{cesare2015multi, scherer2022resilient} maximize coverage but are brittle under failure—a mid-mission failure can result in total data loss. Some approaches predict teammate positions for implicit coordination \cite{schack2024sound}, but do not directly optimize relay-to-base delivery. Others trigger relay based on fixed information thresholds \cite{clark2021queue}. Since all these fixed-criterion approaches ignore accumulated information, exploration potential, and how relay costs scale with environment size, they lack adaptivity to mission conditions. In contrast, our approach makes adaptive online relay decisions based on predicted future information delivery rates.

A related thread of research focuses on \textit{rendezvous-based inter-robot information sharing} in unknown environments \cite{song2024multi, meghjani2011combining}. These works primarily employ methods where robots meet at predefined times or locations to exchange maps or plans, often treating base station reporting as secondary or leaving it unmodeled. Our work instead prioritizes base station relay as the primary objective, enabling opportunistic inter-robot sharing while maximizing information successfully delivered to the base station within a given time limit.

\begin{figure*}[t!]
    \centering
\includegraphics[width=0.97\linewidth,trim=3mm 38mm 1mm 5mm, clip]{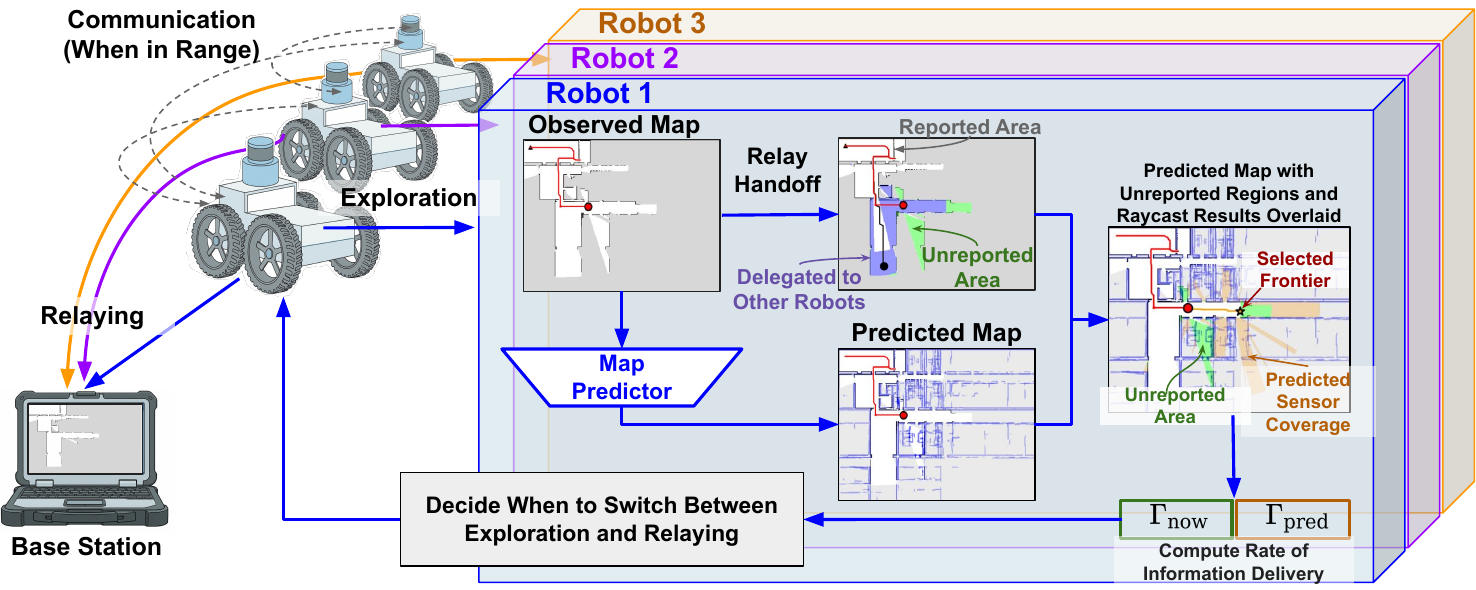}
    \caption{Overview of \textbf{\texttt{PRoID}}: robots explore unknown environments and share map data when in range. Each robot tracks its unique unreported information and continuously evaluates whether to continue exploring or relay to the base, based on comparing current and predicted future rates of information delivery.}
    \label{Fig2}
\end{figure*}

\subsection{Map Prediction for Exploration}
Map prediction has emerged as a powerful tool for improving exploration by reasoning about unseen areas \cite{shrestha2019learned, ramakrishnan2020occupancy, georgakis2022uncertainty, ho2025mapex, baek2025pipe}. The key insight of these works is to train neural networks on 2D floorplan data, enabling prediction of occupancy and structure beyond observed regions given only the observed area. Planners then reason about these predictions. Existing approaches diverge into two main directions: some focus on identifying regions with high visible utility in the predicted space \cite{shrestha2019learned}, while others aim to reduce prediction uncertainty itself \cite{ramakrishnan2020occupancy, georgakis2022uncertainty}. A third class of work reasons jointly to maximize both uncertainty reduction and potential sensor coverage \cite{ho2025mapex}. Recent advances extend this further by performing raycasting along the entire path to the robot's next waypoint, allowing systems to estimate future sensor coverage gains \cite{baek2025pipe}. Our work directly leverages this line of research: we exploit predicted maps and employ raycasting to estimate future sensor coverage. However, rather than focusing solely on frontier selection, we use estimated sensor coverage as an adaptive criterion to determine when a robot should return to the base station for information relay.

\section{Problem Formulation}
\label{problem}
\subsection{Multi-Robot Exploration and Relaying (MRER)}
\label{problem-MRER}
We formulate a decentralized multi-robot exploration and relaying problem in a 2D indoor environment $\mathcal{E} \subset \mathbb{R}^2$, where $n$ robots $\{r_1, \dots, r_n\}$ explore unknown regions and relay collected information to a base station. Each robot $r_i$ is localized by its 2D position $\mathbf{x}_{i,t} = (x_{i,t}, y_{i,t})$ in a shared global frame; all robots start from a common initial location $\mathbf{x}_0$ at $t=0$, and the base station $B$ remains fixed at $\mathbf{x}_b = \mathbf{x}_0$. Each robot is equipped with a LiDAR with fixed sensing range $l$ and maintains a local occupancy grid map $O_{i,t}$. Each robot travels at constant speed $v$; we assume that travel time between any two positions can be estimated via A* path length divided by $v$.

We model communication constraints via a binary function $C : \mathcal{E} \times \mathcal{E} \rightarrow \{0,1\}$, where $C(\mathbf{x}_i, \mathbf{x}_j) = 1$ indicates that robots at positions $\mathbf{x}_i, \mathbf{x}_j$ can exchange information, and $0$ otherwise. Specifically, communication is permitted when the Euclidean distance between two positions falls below a threshold $d$:
\begin{equation}
    C(\mathbf{x}_i, \mathbf{x}_j) = \begin{cases}
1 & \text{if}~~\|\mathbf{x}_i-\mathbf{x}_j\| < d\\
0 & \text{otherwise}
\end{cases}
\end{equation}
When two robots can communicate ($C(\mathbf{x}_{i,t}, \mathbf{x}_{j,t}) = 1$), they fuse their local occupancy maps via set union to produce a combined map. The same communication model governs information exchange between a robot and the base station $B$: a robot must come within communication range to transmit its collected data. Upon doing so, it not only uploads its local map but also receives the current global map maintained at $B$. The base station aggregates all received local maps into a global map $O_b(t)$. The full information-sharing protocol is described in Sec.~\ref{baseline}.

The objective of the decentralized multi-robot exploration and relaying problem is to generate trajectories $\zeta = \{\zeta_1,..\zeta_n\}$ that maximize the collective understanding of the environment at the base station.
Specifically, we seek to maximize information accumulated in the base station map $O_b(T)$:
$$\arg \underset{\zeta}{\max} \mathbb{I}(O_b(T)),$$
where $\mathbb{I}(\cdot)$ denotes an information metric of the map (e.g., coverage), and $T$ is the mission horizon. Note that while deferring all relay to the mission end maximizes exploration time, intermediate relays allow robots to receive updated global maps from the base station, enabling better coordination and reducing redundant coverage upon resuming exploration.

\subsection{MRER under Probability of Failure}
\label{problem-MRER-failure}
We extend MRER by incorporating robot failure: a robot may fail at any time during exploration or relaying, permanently ceasing all operation. Any information collected but not yet transmitted to the base station is then irrevocably lost.

To model the probability of failure, we assume that the lifetime of each robot follows a Weibull distribution \cite{hallinan1993review}. The cumulative probability of failure by time $t$ is given by
\begin{equation}
    F(t) = 1 - \exp\left( - \left( \frac{t}{\lambda} \right)^k \right),
    \label{weibull-cdf}
\end{equation}
where $\lambda > 0$ is the failure timescale parameter and $k > 0$ governs whether failure risk accelerates over time. The hazard rate at time $t$ is
\begin{equation}
    h(t) = \frac{k}{\lambda} \left( \frac{t}{\lambda} \right)^{k-1}.
    \label{hazard}
\end{equation}

We assume that each robot has access to the parameters of this failure model and can therefore incorporate the probability of failure into its planning and decision-making process.

\section{Method}

Our approach builds on a frontier-based multi-robot exploration system with map sharing and relay handoff (Sec.~\ref{baseline}). On top of this, we introduce \textbf{\texttt{PRoID}} (Predicted Rate of Information Delivery), an adaptive relay criterion that compares each robot's current rate of information delivery against its predicted future rate along its planned path (Sec.~\ref{proid}, Fig.~\ref{Fig2}). \textbf{\texttt{PRoID-Safe}} extends this with a survival-weighted criterion that biases relay decisions toward earlier return as failure risk grows (Sec.~\ref{proid_safe}).

\subsection{Preliminary: Baseline System Components}
\label{baseline}
The following components are shared across all methods, including our baselines.

\noindent{\textbf{Frontier-based Exploration Framework:}} Each robot $r_i$ follows a frontier-based exploration strategy:
\[
        \mathcal{F}_{i,t} \leftarrow \textsc{Extract}(O_{i,t})\]
\vspace{-6mm}
\[
        f_{i,t}^* =
\arg\max_{f \in \mathcal{F}_{i,t}} 
\textsc{Score}(f).
\]
Frontier cells are defined as the boundary between known and unknown regions in the occupancy grid map and are clustered into candidate frontier points $\mathcal{F}_{i,t}$. The robot selects the next exploration target by maximizing a scoring function, which may prioritize proximity (nearest-frontier) or expected information gain, depending on the configuration. We use the algorithm of \cite{baek2025pipe} as a frontier scoring function.

\noindent{\textbf{Information Sharing and Coordination:}}
If robots $r_i$ and $r_j$ satisfy 
$C(\mathbf{x}_{i,t}, \mathbf{x}_{j,t}) = 1$, they share their maps, and update them by fusing them. 
\[
\tilde{O}_{ij,t} = O_{i,t} \cup O_{j,t},
\]
\vspace{-6mm}
\[
O_{i,t}, O_{j,t} \leftarrow \tilde{O}_{ij,t}.
\]
The fused map is broadcast back to both robots, ensuring a consistent shared view of the environment.
In addition, robots share their past trajectories $\zeta_{i,t}$ and current plans $\pi_{i,t}$. Let
\[
 \mathcal{N}_{i,t} =
\bigcup_{j \neq i}
\left(
\zeta_{j,t} \cup \pi_{j,t}
\right)
\]
denote the set of spatial commitments of other robots. Robot $r_i$ penalizes frontier candidates that are within distance $\epsilon$ of any element in $\mathcal{N}_{i,t}$ by a large value $\gamma$, thereby discouraging redundant exploration of already-committed regions. Formally:
\[
\textsc{Score}(f) = \textsc{Score}(f) -\gamma
\quad
\text{if}
\quad
\exists p' \in \mathcal{N}_{i,t}
\text{ s.t. }
\|f - p' \| \le \epsilon.
\]
We ablate the effect of trajectory and plan sharing in Sec.~\ref{experiments}.

\noindent{\textbf{Relay Handoff:}} Additionally, we consider a relay handoff mechanism to improve efficiency. Suppose that robot $r_i$ is currently returning to the base station to relay its collected information and encounters another robot $r_j$ within communication range. If the distance between $r_j$ and the base station is smaller than that between $r_i$ and the base station, i.e.,
\[
\text{if}\quad C(\mathbf{x}_{i,t},\mathbf{x}_{j,t})=1 \quad \& \quad 
\|\mathbf{x}_{j,t} - \mathbf{x}_b\| < \|\mathbf{x}_{i,t} - \mathbf{x}_b\|,
\]
then robot $r_i$ transfers all its accumulated information to $r_j$ and delegates the relaying task to $r_j$. After the handoff, robot $r_i$ immediately resumes exploration, thereby reducing idle travel time and enabling more efficient task allocation within the team. We ablate this mechanism in Sec.~\ref{experiments} to quantify its contribution to overall performance.

\noindent{\textbf{Final Return:}} Regardless of the relaying strategy, all robots enforce a mission-end guarantee: each robot continuously tracks the remaining mission time and initiates a final relay trip once the remaining time equals the estimated travel time back to the base station, ensuring all accumulated data is delivered before the mission concludes.

\begin{figure}[t!]
    \centering
\includegraphics[width=0.9\linewidth, trim=2mm 30mm 171mm 2mm, clip]{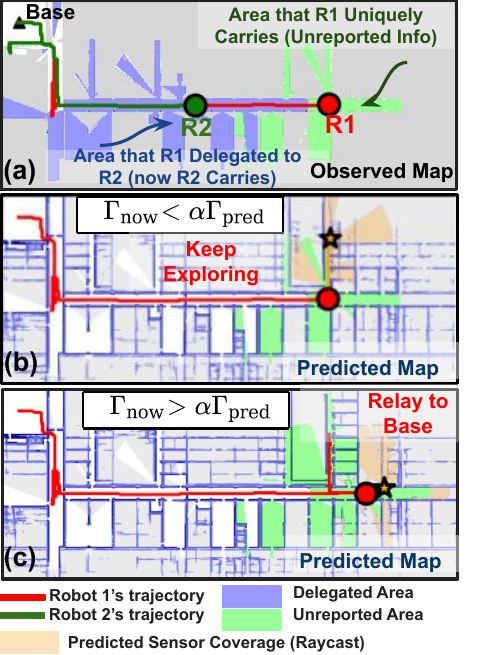}
    \caption{(a) Robot 1 shares its map with Robot 2, delegating overlapping coverage (blue) and retaining its unique unreported area (green). (b) Robot 1 estimates predicted sensor coverage along its planned path (orange) and compares current vs.\ predicted rate of information delivery--deciding to continue exploring. (c) After further exploration, the current rate exceeds the predicted rate, triggering relay to the base station.}
    \label{Fig3}
\end{figure}

\subsection{Predicting Rate of Information Delivery (\textbf{\texttt{PRoID}})}
\label{proid}
We introduce \textbf{\texttt{PRoID}} in two steps: we first define the Rate of Information Delivery (RoID), then present how to predict it to drive adaptive relay decisions.

\noindent{\textbf{Rate of Information Delivery (RoID):}} While a robot's information gain naturally grows as it explores unknown regions, mere accumulation of data does not equate to utility; the value of collected information must be weighed against the time required to deliver it to the base station. 

We define RoID as the ratio of novel, reportable information to the estimated travel time back to the base station. From the base station's perspective, previously reported data is redundant and provides no additional utility. Furthermore, in our multi-robot framework, any region already delegated to another robot through relay handoff (as discussed in Sec.~\ref{baseline}) must be excluded from the current robot's information payload. Thus, the effective information payload  $I_{\text{unreported}}$ is computed by excluding the already-reported cells $O_{\text{reported},i,t}$ and the delegated regions $O_{\text{delegated},i,t}$:
\begin{equation}
    I_{\text{unreported}} = \mathbb{I}(O_{i,t} \setminus (O_{\text{reported},i,t} \cup O_{\text{delegated},i,t}))
\end{equation}
The current Rate of Information Delivery, $\Gamma_\text{now}$, is defined as: 
\begin{equation}
    \Gamma_\text{now} = \frac{I_{\text{unreported}}}{t_{\text{cur}\rightarrow \text{base}}}
\end{equation}
where $t_{\text{cur}\rightarrow \text{base}}$ denotes the estimated time required for the robot to travel from its current position to the base station.

\noindent{\textbf{Predicted Rate of Information Delivery (\textbf{\texttt{PRoID}})}}: 
Recent work in indoor exploration \cite{ho2025mapex, baek2025pipe} has leveraged learned map prediction models to infer plausible structures beyond the current observed area. In this work, we adopt the approach of~\cite{baek2025pipe}, which enables fast estimation of potential cumulative sensor coverage over predicted maps. Specifically, each robot maintains a pretrained map predictor $\mathcal{G}$ that infers a completed map from its local occupancy grid $O_{i,t}$:
\[
M_{i,t} \leftarrow \mathcal{G}(O_{i,t}),
\]
where $M_{i,t}$ denotes the predicted map conditioned on $O_{i,t}$.

For a selected frontier $f^*$, let $\pi_{i,t}(f^*)$ denote the planned A* path from the robot's current position to $f^*$. To estimate the anticipated information gain along this path, we simulate sensor observations over the predicted map $M_{i,t}$. For each point $p \in \pi_{i,t}(f^*)$, we perform ray casting over $M_{i,t}$, yielding a set of visible cells:
\[
\rho(p) \leftarrow \textsc{RayCast}(p, M_{i,t}), \quad \forall p \in \pi_{i,t}(f^*).
\]

The overall visibility mask $\nu$ along the path is defined as the union of visible cells obtained from all ray-casting operations:
\[
\nu \leftarrow \textsc{FloodFill}\!\left( \bigcup_{p \in \pi_{i,t}(f^*)} \rho(p) \right).
\]

The predicted information gain is then defined as
\[
I_{\text{pred}} = \mathbb{I}(\nu),
\]
where $\mathbb{I}(\cdot)$ measures the number of newly observable cells.

The Predicted Rate of Information Delivery is defined as
\begin{equation}
\label{eqn-proid}
    \Gamma_{\text{pred}} =
\frac{
I_{\text{unreported}} + \mathbb{E}[I_{\text{pred}}]
}{
t_{\text{cur} \rightarrow \text{front}} +
t_{\text{front} \rightarrow \text{base}}
},
\end{equation}
where $t_{\text{cur} \rightarrow \text{front}}$ denotes the estimated travel time from the current position to the selected frontier, and $t_{\text{front} \rightarrow \text{base}}$ denotes the estimated return time from the frontier to the base station.

Since $I_{\text{pred}}$ depends on uncertain map completion, we compute its expectation using an ensemble of predicted maps and probabilistic ray casting, following~\cite{ho2025mapex, baek2025pipe}.

\noindent{\textbf{Relay Strategy using \texttt{PRoID}:}} We compare the current Rate of Information Delivery (RoID), $\Gamma_{\text{now}}$, with the Predicted Rate of Information Delivery (\textbf{\texttt{PRoID}}), $\Gamma_{\text{pred}}$, to determine when a robot should return to the base station for data relay.

While $\Gamma_{\text{now}}$ measures the efficiency of immediate return, $\Gamma_{\text{pred}}$ measures the expected efficiency of exploring first and then returning. To balance exploration and relaying efficiency, we enforce the following decision rule:
\begin{equation}
\Gamma_{\text{now}} > \alpha \, \Gamma_{\text{pred}},
\label{criterion}
\end{equation}
under which the robot terminates exploration and initiates relay to the base station.

The parameter $\alpha \geq 1$ introduces a conservative bias toward exploration. It prevents premature returns caused by short-term fluctuations in predicted gain and accounts for uncertainty in map prediction and travel time estimation. This threshold-based comparison can be interpreted as a myopic one-step lookahead policy that selects between immediate relay and frontier-conditioned exploration based on normalized information return. In Sec.~\ref{experiments}, we ablate the effect of $\alpha$.

Fig.~\ref{Fig3} illustrates \textbf{\texttt{PRoID}} in action: a robot delegates observed area to a teammate, estimates predicted future coverage along its planned path, and triggers relay once the current delivery rate exceeds the predicted rate.

\subsection{Failure-Aware Relaying Strategy (\textbf{\texttt{PRoID-Safe}})}
\label{proid_safe}
As defined in Sec.~\ref{problem-MRER-failure}, we consider a Weibull-distributed failure model, where the instantaneous probability of failure at time $t$ is governed by the hazard function (Eq.~\ref{hazard}) and the cumulative probability that a robot has failed by time $t$ is given by $F(t)$ (Eq.~\ref{weibull-cdf}). Then the corresponding survival function is
\begin{equation}
    S(t) = 1 - F(t) = \exp\!\left(-\left(\frac{t}{\lambda}\right)^k\right),
    \label{survival}
\end{equation}
which represents the probability that the robot remains operational up to time $t$.

If the robot immediately returns to the base station, and the travel time to the base is $t_{\text{cur}\rightarrow\text{base}}$, the probability that the robot survives until completing the relay is
\[
S_{\text{now}} = S\!\left(t + t_{\text{cur}\rightarrow\text{base}}\right).
\]

Alternatively, if the robot first travels to the next selected frontier, requiring time $t_{\text{cur}\rightarrow\text{front}}$, and then returns from the frontier to the base station in time $t_{\text{front}\rightarrow\text{base}}$, the survival probability becomes
\[
S_{\text{pred}} = S\!\left(t + t_{\text{cur}\rightarrow\text{front}} + t_{\text{front}\rightarrow\text{base}}\right).
\]

We propose \textbf{\texttt{PRoID-Safe}}, a survival-weighted relaying strategy that incorporates failure risk into the decision process. Rather than comparing raw rates of information delivery, the robot evaluates the \emph{expected} rate of successfully delivered information by weighting each rate by its corresponding survival probability. Formally, the robot initiates relay when
\begin{equation}
    \Gamma_{\text{now}} \, S_{\text{now}}
>
\alpha \, \Gamma_{\text{pred}} \, S_{\text{pred}},
\end{equation}
where $\Gamma_{\text{now}}$ and $\Gamma_{\text{pred}}$ denote the instantaneous and predicted rates of information delivery, as defined in Sec.~\ref{proid}.

This criterion can be interpreted as a risk-adjusted variant of Eq.~\ref{criterion}. By scaling each rate with the probability of survival until delivery, the strategy accounts for the possibility that information collected during extended exploration may never reach the base due to failure. 

\section{Experiments}
\label{experiments}

We evaluate \textbf{\texttt{PRoID}} and \textbf{\texttt{PRoID-Safe}} across varying team sizes and failure conditions to assess their effectiveness against fixed-schedule baselines. Experiments cover both no-failure and failure-prone scenarios, and include ablations on system components and hyperparameters.

\subsection{Experimental Setup (Dataset and Metrics)}
\begin{figure}[t!]
    \centering
\includegraphics[width=1.0\linewidth]{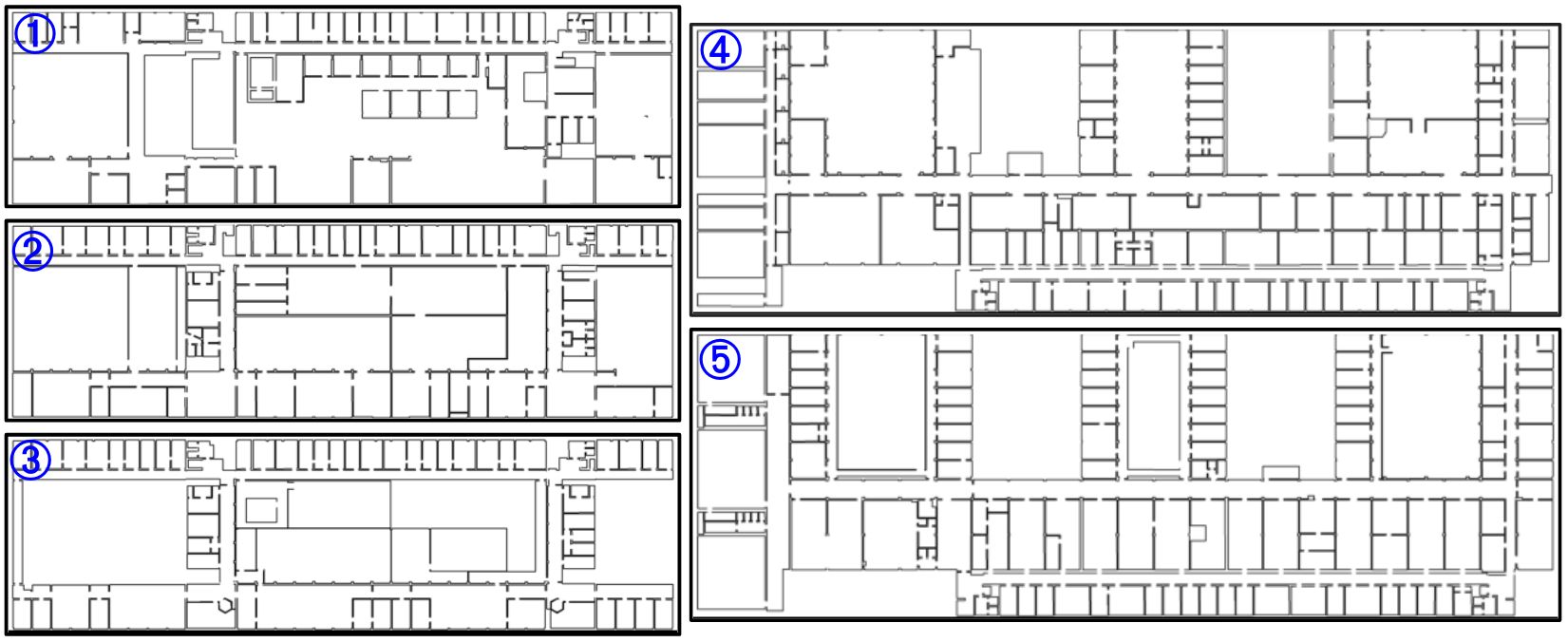}
    \caption{Five Test Maps from KTH Indoor Floorplan Dataset \cite{aydemir2012can}.}
    \label{kth-maps}
\end{figure}

We built a multi-robot indoor simulator on the KTH floor plan dataset~\cite{aydemir2012can}, split into train and test partitions. The training set fine-tunes an image inpainting network~\cite{suvorov2022resolution} as the map predictor; all experiments use the test set. The simulator uses a pixel-to-meter ratio of $10{:}1$ and raycast range $l=20m$. We set the communication range $d=10m$. We evaluate across (i) robot count $n \in \{2,3,4,5\}$, (ii) five test maps (three of $60m \times 205m$ and two of $88m \times 265m$) as shown in Fig.~\ref{kth-maps}, and (iii) five random starting locations per map, giving $25$ unique initial configurations per robot count. For failure scenarios, we fix $k=1.5$ and evaluate two values of $\lambda \in \{1100,900\}$. We set the mission horizon $T=1000$. We report the coverage ratio of the base station map at mission end, $\mathbb{I}(O_b(T)) / |\mathcal{E}|$, which directly measures the objective defined in Sec.~\ref{problem-MRER}.

\subsection{Baselines and Ablation Study}
For baselines, we compare our method against two fixed-schedule relay strategies:
\begin{itemize}
    \item \textbf{Periodic Relay:} Each robot relays to the base station at a fixed interval $P$, then resumes exploration. We evaluate $P \in \{100, 200, 300\}$ timesteps.
    \item \textbf{Final Relay Only:} Each robot explores until the deadline and returns to the base station only once. This maximizes exploration time but is brittle under failure.
\end{itemize}

We further conduct ablation studies on a representative subset ($n=3$ robots, three starting poses per map):
\begin{itemize}
    \item \textbf{Relay Handoff:} Full system vs.\ Relay Handoff disabled
    \item \textbf{Trajectory and Plan Sharing:} With vs.\ without inter-robot commitment sharing
    \item \textbf{Effect of $\alpha$:} $\alpha \in \{1.0, 1.2, 1.5, 2.0\}$ for \textbf{\texttt{PRoID}}
\end{itemize}

\begin{figure}[t!]
    \centering
\includegraphics[width=1.0\linewidth, trim=0mm 50mm 98mm 0mm, clip]{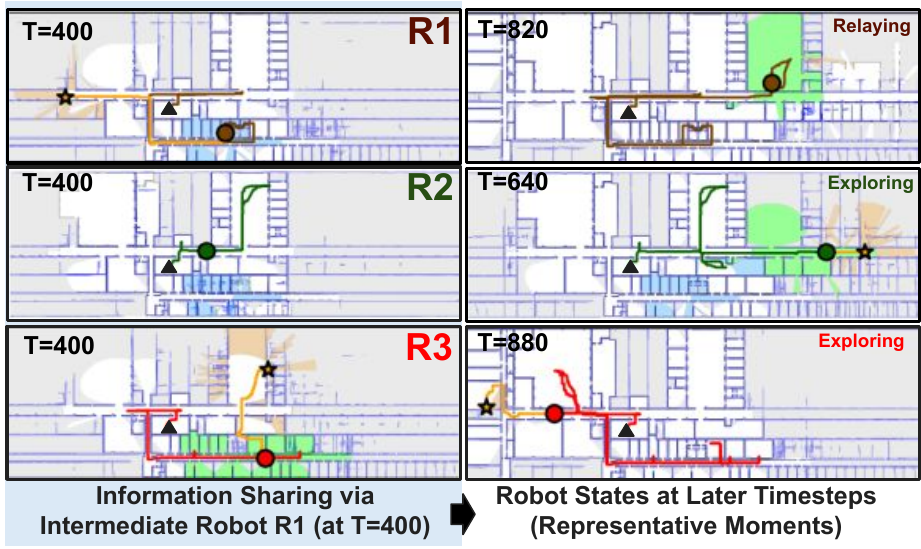}
    \caption{\textbf{(Left)} At $T=400$, R1 delegates observed area (blue) to R3. R2, out of range of R3, still learns these areas are delegated via R1. \textbf{(Right)} Each robot compares unique unreported area (green) against predicted future coverage (orange). R1 relays (small predicted coverage); R2 explores; R3, having fully delivered all data, continues exploring.}
    \label{Fig6}
\end{figure}

\subsection{Results}
\begin{figure*}[t!]
\centering
\includegraphics[width=0.95\linewidth]{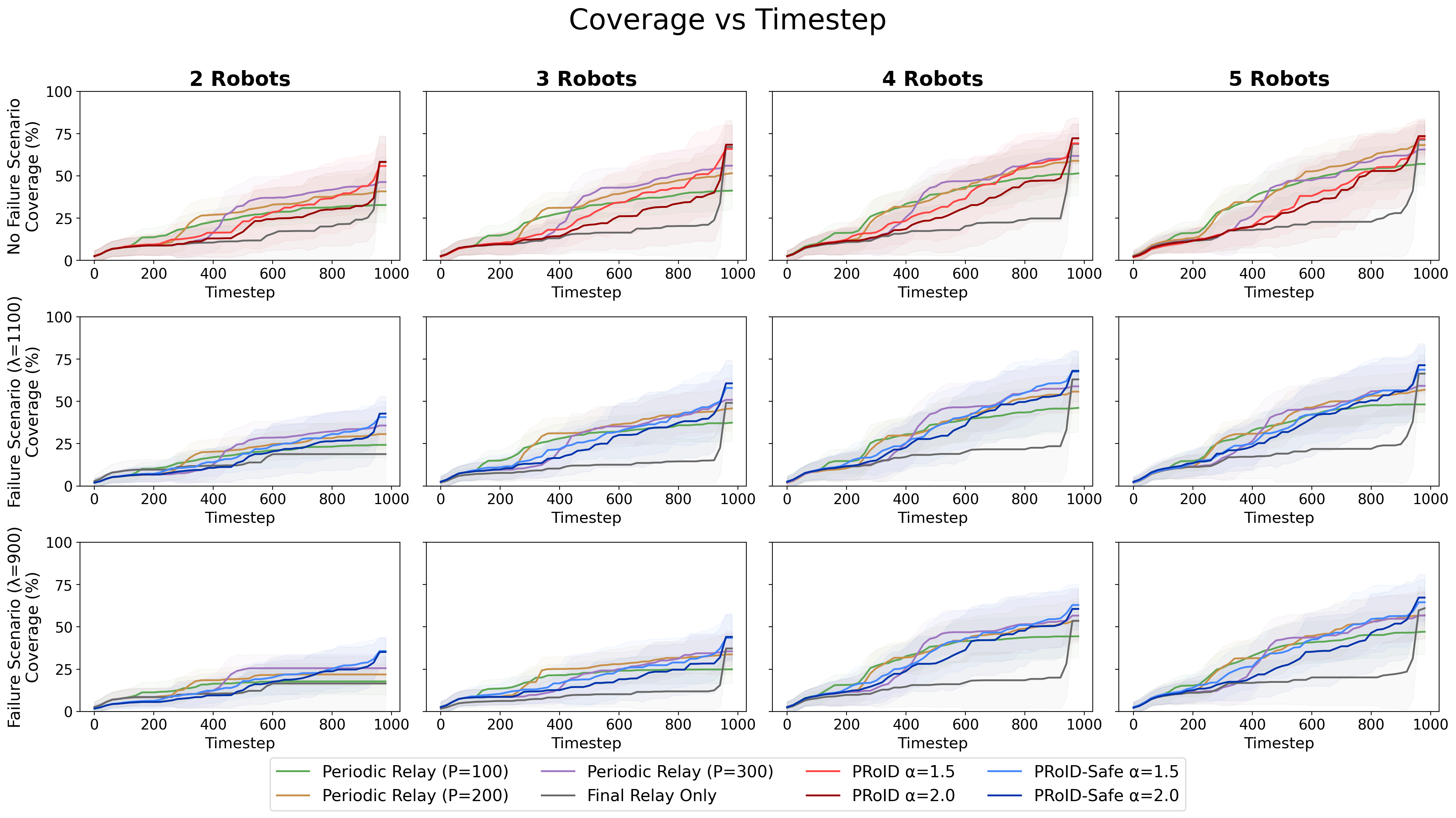}
\caption{Quantitative comparison of \textbf{\texttt{PRoID}} against fixed-schedule baselines in the no-failure scenario, and \textbf{\texttt{PRoID-Safe}} in failure-prone scenarios ($\lambda \in \{1100, 900\}$). Our methods consistently outperform baselines in total information delivered to the base station.}
\label{graph-figure}
\end{figure*}

\begin{table*}[t]
\centering
\caption{Coverage ratio (\%) at mission end, averaged over all test configurations. Best results per column in \textbf{bold}.}
\label{tab:results}
\small
\begin{tabular}{lcccccccccccc}
\toprule
& \multicolumn{4}{c}{No Failure Scenario} & \multicolumn{4}{c}{Failure Scenario ($\lambda=1100$)} & \multicolumn{4}{c}{Failure Scenario ($\lambda=900$)} \\
\cmidrule(lr){2-5}\cmidrule(lr){6-9}\cmidrule(lr){10-13}
Method & $n{=}2$ & $n{=}3$ & $n{=}4$ & $n{=}5$ & $n{=}2$ & $n{=}3$ & $n{=}4$ & $n{=}5$ & $n{=}2$ & $n{=}3$ & $n{=}4$ & $n{=}5$ \\
\midrule
Periodic Relay ($P=100$)    & 32.8 & 41.3 & 51.5 & 57.1 & 24.3 & 37.4 & 46.2 & 48.2 & 17.7 & 24.9 & 44.4 & 47.1 \\
Periodic Relay ($P=200$)    & 40.9 & 51.6 & 59.0 & 68.2 & 30.7 & 45.8 & 55.8 & 56.8 & 21.9 & 33.7 & 53.7 & 56.9 \\
Periodic Relay ($P=300$)    & 46.4 & 56.0 & 61.8 & 65.7 & 35.7 & 50.9 & 58.9 & 59.2 & 25.6 & 35.5 & 56.6 & 56.7 \\
Final Relay Only            & \textbf{58.4} & 67.1 & 68.9 & 71.4 & 18.8 & 49.1 & 63.0 & 66.4 & 16.5 & 37.2 & 53.6 & 61.0 \\
\midrule
\textbf{\texttt{PRoID}} (Ours)      & 58.2 & \textbf{68.4} & \textbf{72.3} & \textbf{73.5} & 34.0 & 57.4 & 65.1 & 68.2 & 32.3 & 42.0 & 58.3 & 60.1 \\
\textbf{\texttt{PRoID-Safe}} (Ours) & -- & -- & -- & -- & \textbf{42.8} & \textbf{60.7} & \textbf{68.1} & \textbf{71.4} & \textbf{35.7} & \textbf{44.1} & \textbf{62.9} & \textbf{67.3} \\
\bottomrule
\end{tabular}
\end{table*}

As shown in Table~\ref{tab:results} and Fig.~\ref{graph-figure}, in the no-failure scenario, \textbf{\texttt{PRoID}} outperforms the best periodic baseline ($P=300$) by 11.8, 12.4, 10.5, and 7.8 percentage points at $n=2,3,4,5$ respectively, and outperforms Final Relay Only by 1.3, 3.4, and 2.1 percentage points at $n=3,4,5$.
At $n=2$, Final Relay Only is marginally competitive (58.4\% vs.\ 58.2\%): with only two robots, a relay trip reduces the active exploration force by half, limiting the benefit of adaptive scheduling.
As team size grows, this overhead diminishes and \textbf{\texttt{PRoID}}'s relay handoff and coordination advantages become increasingly pronounced.

In failure scenarios, \textbf{\texttt{PRoID-Safe}} consistently outperforms all baselines.
Against the best periodic baseline ($P=300$), \textbf{\texttt{PRoID-Safe}} achieves gains of 9.8 and 12.2 percentage points at $n=3$ and $n=5$ under $\lambda=1100$, and 8.6 and 10.6 percentage points under the harsher $\lambda=900$ condition.
Final Relay Only degrades most severely: at $n=2$, coverage drops to just 18.8\% ($\lambda=1100$) and 16.5\% ($\lambda=900$)---compared to \textbf{\texttt{PRoID-Safe}}'s 42.8\% and 35.7\%---as robots frequently fail before completing the single return trip.
While periodic relay baselines are more resilient since data is delivered throughout the mission, their fixed schedules cannot adapt to growing failure risk, and \textbf{\texttt{PRoID-Safe}} maintains consistent advantages over them across all team sizes and failure rates.

Ablation results are reported in Table~\ref{tab:ablation} ($n=3$, no-failure scenario).
Disabling relay handoff and trajectory/plan sharing both reduce performance, confirming that each coordination component contributes to performance.
For the effect of $\alpha$, lower values ($\alpha=1.0, 1.2$) trigger relay too early and reduce overall coverage, while $\alpha=2.0$ provides the best trade-off.

Fig.~\ref{Fig6} illustrates multi-robot coordination behavior in \textbf{\texttt{PRoID}}.
At $T$=400, R1 acts as an information hub: it delegates its observed area to R3 (shown in blue on R1's map), while R2--out of direct communication range of R3--correctly learns through R1 that these areas are already delegated.
This indirect propagation extends the effective information-sharing range of the team beyond pairwise communication.
In the right panels, each robot independently compares its unique unreported area (green) against predicted future sensor coverage (orange).
R1, whose predicted coverage is small relative to its unreported data, triggers relay to the base station.
R2 continues exploring, maintaining a healthy balance between unreported and predicted coverage.
R3, having already delivered all its data, carries no unreported area and continues exploring freely--the ideal outcome of timely relaying.

\begin{figure}[t!]
    \centering
\includegraphics[width=0.97\linewidth, trim=2mm 50mm 150mm 0mm, clip]{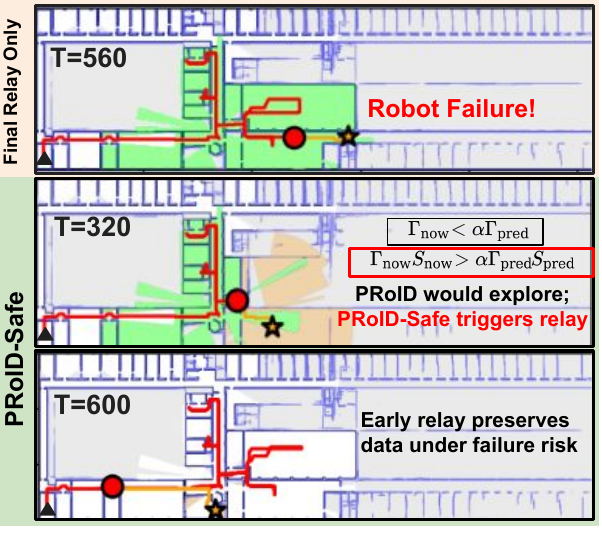}
    \caption{\textbf{(Top)} Final Relay Only robot fails at $T=560$ before returning, losing all data. \textbf{(Middle)} At $T=320$, \textbf{\texttt{PRoID-Safe}} triggers relay despite $\Gamma_\text{now} < \alpha\Gamma_\text{pred}$, as survival probability shifts the criterion ($\Gamma_\text{now} S_\text{now} > \alpha\Gamma_\text{pred} S_\text{pred}$). \textbf{(Bottom)} Data already relayed; robot continues exploring safely.}
    \label{Fig7}
\end{figure}

Fig.~\ref{Fig7} illustrates the benefit of \textbf{\texttt{PRoID-Safe}} under robot failure.
In the top panel, a robot operating under Final Relay Only fails at $T$=560 before completing its return trip, losing all accumulated data (green) irrecoverably.
In contrast, \textbf{\texttt{PRoID-Safe}} biases relay decisions toward earlier return by incorporating survival probability: at $T$=320, although the standard \textbf{\texttt{PRoID}} criterion would indicate continued exploration ($\Gamma_\text{now} < \alpha \Gamma_\text{pred}$), the failure-aware criterion triggers relay ($\Gamma_\text{now} S_\text{now} > \alpha \Gamma_\text{pred} S_\text{pred}$), reflecting growing failure risk.
By $T$=600, the robot has already relayed its data and continues operating, preserving information that would otherwise be lost.

\begin{table}[t]
\centering
\caption{Ablation study. Coverage ratio (\%) at $n=3$, averaged over test configurations (no-failure scenario).}
\label{tab:ablation}
\small
\begin{tabular}{lc}
\toprule
Configuration & Coverage (\%) \\
\midrule
\multicolumn{2}{l}{\textit{System Components}} \\
\textbf{\texttt{PRoID}} Full System (Ours)     & 67.9 \\
~~w/o Relay Handoff                            & 59.8 \\
~~w/o Traj.\ \& Plan Sharing                  & 53.5 \\
\midrule
\multicolumn{2}{l}{\textit{Effect of $\alpha$ — \textbf{\texttt{PRoID}}}} \\
$\alpha = 1.0$ & 49.0 \\
$\alpha = 1.2$ & 61.1 \\
$\alpha = 1.5$ & 66.3 \\
$\alpha = 2.0$ & 67.9 \\
\bottomrule
\end{tabular}
\end{table}

\section{Conclusions}
We presented \textbf{\texttt{PRoID}}, an adaptive relay criterion for multi-robot exploration that compares each robot's current rate of information delivery against its predicted future rate along its planned path. By accounting for unique unreported data, teammate coordination, and learned map prediction, \textbf{\texttt{PRoID}} enables decentralized relay decisions without fixed schedules or pre-planned rendezvous. Its failure-aware extension, \textbf{\texttt{PRoID-Safe}}, incorporates survival probability to bias relay decisions under increasing failure risk. Experiments on real-world indoor floor plans demonstrate consistent improvements over fixed-schedule baselines across team sizes and failure conditions. Future work includes integrating learned navigation policies, leveraging map predictions to estimate teammate positions, and validating the approach on real hardware.

\ifdefined\conferenceversion
\else
\appendix
\renewcommand{\thesection}{A\arabic{section}}
\setcounter{table}{0}
\renewcommand{\thetable}{A.\arabic{table}}
\renewcommand{\thealgorithm}{\arabic{algorithm}}

\section*{\centering A1. Algorithmic Details}
In Alg.~\ref{alg:proid}, we detail the algorithmic procedures of \textbf{\texttt{PRoID}} and \textbf{\texttt{PRoID-Safe}}.

\begin{algorithm}[t]
\caption{\textbf{\texttt{PRoID}} and \textbf{\texttt{PRoID-Safe}}} \label{alg:proid} 
\begin{algorithmic}[0]
\Require Robots $R_i$ ($i=1,2,..,N$), Base Station $B$, Budget $T$ \\
$\triangleright$ Algorithm runs on each $R_i$. Starts with \texttt{Explore} mode.
\State \textbf{for} $t=1$ to $T$:
\State \hspace{1.2mm} $O_{i,t} \leftarrow$ \textsc{Raycast}($O_{i,t-1}, \mathbf{x}_{i,t}, l$) \Comment{Observe}
\State \hspace{1.2mm} \textbf{for} $\forall j \in \{1,2,..,N\}$ s.t. $i \neq j$:
\State \hspace{2.4mm} \textbf{if} $C(\mathbf{x}_{i,t},\mathbf{x}_{j,t})=1$: \Comment{Communication Available}
\State \hspace{3.6mm} $O_{i,t} \leftarrow O_{i,t} \cup O_{j,t}$ \Comment{Combine Observed Maps}
\State \hspace{3.6mm} $\mathcal{N}_{i,t} =\bigcup_{j \neq i}\left(\zeta_{j,t} \cup \pi_{j,t} \right)$ \Comment{Share Trajectories \& Plans}
\State \hspace{3.6mm} \textbf{if} $R_i.\text{mode}==\texttt{Relay}$ \textbf{and} $\|\mathbf{x}_{j,t} - \mathbf{x}_b\| < \|\mathbf{x}_{i,t} - \mathbf{x}_b\|$:
\State \hspace{4.8mm} Delegate information from $R_i$ to $R_j$ \Comment{Relay Handoff}
\State \hspace{4.8mm} $R_j.\text{mode}\leftarrow\texttt{Relay}$, $R_i.\text{mode}\leftarrow\texttt{Explore}$
\State \hspace{1.2mm} \textbf{if} $T - t \leq t_{\mathbf{x}_{i,t}\rightarrow\mathbf{x}_b}$: \Comment{Final Return}
\State \hspace{2.4mm} $R_i.\text{mode} \leftarrow \texttt{Relay}$
\State \hspace{1.2mm} \textbf{if} $R_i.\text{mode}==\texttt{Explore}$:
\State \hspace{2.4mm} \textbf{if} robot needs to select a new waypoint:
\State \hspace{3.6mm} $\mathcal{F}_{i,t} \leftarrow$ \textsc{Extract}($O_{i,t}$) \Comment{Frontiers}
\State \hspace{3.6mm} $M_{i,t} \leftarrow \mathcal{G}(O_{i,t})$ \Comment{Generate Predictions}
\State \hspace{3.6mm} \textbf{for} frontier $f$ in $\mathcal{F}_{i,t}$:
\State \hspace{4.8mm} score each frontier $f$ using \cite{baek2025pipe}
\State \hspace{4.8mm} $f.\text{score} = f.\text{score} - \gamma$ \textbf{if} $\exists p' \in \mathcal{N}_{i,t}
\text{ s.t. }\|f - p'\| \le \epsilon$
\State \hspace{3.6mm} $f^* \leftarrow \underset{f\in \mathcal{F}}{\arg \max}(f.\text{score})$ \Comment{Choose Waypoint}
\State \hspace{2.4mm} Follow $\pi_{i,t}(f^*)$, the A-star path to $f^*$
\State \hspace{2.4mm} $\Gamma_\text{now} \leftarrow I_\text{unreported}/t_{\mathbf{x}_{i,t}\rightarrow\mathbf{x}_b}$ \Comment{Compute RoID}
\State \hspace{2.4mm} $\triangleright$ \textbf{Compute PRoID:}  
\State \hspace{2.4mm} $\rho(p) \leftarrow \textsc{RayCast}(p, M_{i,t}), \quad \forall p \in \pi_{i,t}(f^*)$
\State \hspace{2.4mm} $\nu \leftarrow \textsc{FloodFill} (\bigcup_{p \in \pi_{i,t}(f^*)} \rho(p)) $
\State \hspace{2.4mm} $\Gamma_\text{pred} \leftarrow (I_\text{unreported} + \mathbb{E}[\mathbb{I}(\nu)])/(t_{\mathbf{x}_{i,t} \rightarrow f^*} +
t_{f^* \rightarrow \mathbf{x}_b})$
\State \hspace{2.4mm} \textbf{if} $\Gamma_{\text{now}} > \alpha \Gamma_\text{pred}$: \Comment{\textbf{\texttt{PRoID}}}
\State \hspace{2.4mm} (or \textbf{if} $\Gamma_{\text{now}} S_{\text{now}} > \alpha \Gamma_\text{pred} S_{\text{pred}} $:) \Comment{\textbf{\texttt{PRoID-Safe}}}
\State \hspace{3.6mm} $R_i$.mode $\leftarrow$ \texttt{Relay}
\State \hspace{1.2mm} \textbf{if} $R_i.\text{mode}==\texttt{Relay}$:
\State \hspace{2.4mm} Follow A-star Path to $\mathbf{x}_B$
\State \hspace{2.4mm} \textbf{if} $C(\mathbf{x}_{i,t}, \mathbf{x}_B)=1$: \Comment{Communication with Base}
\State \hspace{3.6mm} Report information to $B$, then $R_i$.mode $\leftarrow$ \texttt{Explore}
\Ensure Final observed map at base station $O_b(T)$
\end{algorithmic}
\end{algorithm}

\section*{\centering A2. Hyperparameter Details}
We provide a list of hyperparameters used in our experiments for reproducibility in Tab.~\ref{tab:proid-hyperparameters}

\begin{table}[h]
\caption{Hyperparameters of \textbf{\texttt{PRoID}} and \textbf{\texttt{PRoID-Safe}}}\label{tab:proid-hyperparameters}
\centering
\begin{tabular}{ll}
\hline
\multicolumn{1}{c}{parameter}              & value                  \\ \hline
$\text{Communication Range}\quad d$               & 10m                    \\
$\text{Raycast Range}\quad l$               & 20m                    \\
Frontier penalty distance: & \\
\quad For trajectories $\epsilon_\text{traj}$ & 5m \\
\quad For plans $\epsilon_\text{plan}$ & 10m \\
Frontier penalty value $\gamma$ & $10^6$ \\
Weibull scale parameter $\lambda$ (characteristic lifetime) & 1100, 900 \\
Weibull shape parameter $k$ (hazard rate growth) & 1.5 \\
\end{tabular}
\end{table}

We further provide a list of hyperparameters used in the baseline frontier-based exploration algorithm~\cite{baek2025pipe} in Tab~\ref{tab:pipe-hyperparameters}.

\begin{table}[h!]
\caption{Hyperparameters of PIPE Planner~\cite{baek2025pipe} (map-prediction-based frontier exploration baseline)}
\label{tab:pipe-hyperparameters}
\centering
\begin{tabular}{ll}
\hline
\multicolumn{1}{c}{Parameter} & Value \\ \hline
Number of map predictors in ensemble & 3 \\
Probabilistic raycast threshold & 0.5 \\
Use distance transform in planning & True \\
Hypothetical laser range & 20m \\
Number of hypothetical lasers & 250 \\
Path sampling interval for raycast & 25 points \\
Minimum frontier region size (pixels) & 10 \\
\end{tabular}
\end{table}







\bibliographystyle{IEEEtran}
\bibliography{IEEEabrv, IEEEexample}

\end{document}